\documentclass[runningheads,a4paper]{llncs}

\usepackage{amssymb}
\usepackage{graphicx}

\usepackage{sidecap}
\usepackage{lipsum,adjustbox}
\usepackage{tikz}
\usepackage{caption, graphicx}
\usepackage{float, amsmath, url}
\usepackage{hyperref, geometry}
\usetikzlibrary{positioning, fit, arrows.meta, shapes, arrows}
\usepackage{enumitem}

\usepackage{url}

\begin{document}

\mainmatter  


\title{Image Captioning using Deep Stacked LSTMs, Contextual Word Embeddings and Data Augmentation}

\titlerunning{Deep Image Captioning with Stacked LSTMs}

%
%
\author{Sulabh Katiyar%
\and Samir Kumar Borgohain}
\authorrunning{Sulabh Katiyar and Samir Kumar Borgohain}

\institute{Department of Computer Science and Engineering,\\
National Institute of Technology, Silchar, India\\}

%
%

\maketitle

\begin{abstract}
Image Captioning, or the automatic generation of descriptions for images, is one of the core problems in Computer Vision and has seen considerable progress using Deep Learning Techniques. We propose to use Inception-ResNet Convolutional Neural Network as encoder to extract features from images, Hierarchical Context based Word Embeddings for word representations and a Deep Stacked Long Short Term Memory network as decoder, in addition to using Image Data Augmentation to avoid over-fitting. For data Augmentation, we use Horizontal and Vertical Flipping in addition to Perspective Transformations on the images. We evaluate our proposed methods with two image captioning frameworks- Encoder-Decoder and Soft Attention. Evaluation on widely used metrics have shown that our approach leads to considerable improvement in model performance.
\end{abstract}

\section{Introduction}
Generation of captions for images is a vital task for scene understanding which is one of the primary aims of Computer Vision. 
But in addition to recognizing objects that appear in the image, the caption generation model must also be powerful enough to recognize the relationships among them and express these relationships in a natural language. And this requirement, which is to be able to mimic the human ability to transform the information from visual domain to the domain of natural languages, is what makes it an important challenge for deep learning research. Over the last few years, the problem has been tackled by using Convolutional Neural Networks (convnets) to extract an image representation and different versions of Recurrent Neural Networks to generate the caption sentences one word at a time.

For extracting the image representations, pretrained Convolutional Neural Networks (CNNs) are used which are first trained on the ImageNet Dataset\cite{imagenet} for the object recognition task. Generally, the image representations used are of two types: (a) the ones which are extracted from the last few fully connected layers of the CNN which distill the information about the salient objects into a fixed length vector such as those used in Vinyals et. al. \cite{vinyals} and Mao et. al. \cite{mao}, and (b) the ones which are extracted from the Convolutional layers of the CNN which produce a set of vectors each of which contain information about an aspect or region of the image such as the those used in Xu et. al. \cite{attendtell}. We use the latter approach as the former condenses the information of an image into a single vector which may lead to loss of information about specific regions or aspects of the image.

To produce the captions the decoder will use the image representations generated by the decoder along with the previous generated words (to be used as contextual information). There are two main ways to use the image representation generated by the CNN: (a) the image representation is mapped to the same vector space as the hidden state representation of the Recurrent Neural Networks (RNN) decoder and used as the initial hidden and cell state of the decoder as proposed in Vinyals et. al. \cite{vinyals} or, (b) the image representation is mapped to the vector space of input word vector representations and then merged with the word embedding representation of the sequence generated until the previous time-step (by concatenating or adding, element-wisely, the two vectors), as proposed in Mao et. al.\cite{mao}. For our work we use the Inception-ResNet-v2 Convolutional Neural Network as proposed in Szegedy et. al.\cite{incepres}.

For producing the captions, we use a Multilayer Long Short Term Memory Network (LSTM) such that input to the network is the input to the first LSTM layer and output of a LSTM layer is used as input for the next LSTM layer. The output of the last LSTM layer is used as input to the softmax activation layer which produces a vector containing probability of occurrence of each word in the vocabulary. We use the description of LSTM as proposed in Hochreiter et. al \cite{lstm}. At each time-step, in addition to the image representation provided as initial hidden and cell states, the words generated till the previous time-step are used as input to the LSTM. The words are passed through an Embedding Layer which produces dense vector representation of the words. We use the Word Embedding framework proposed in Komninos et. al.\cite{komninos}. Henceforth, in this article,  we will refer to these embeddings as Komninos Embeddings.

In automatic image caption generation process, using the image representation generated by the CNN encoder, the decoder will recurrently generate the caption one word at each time step. Precisely, if we have the input image \textit{I} and the target sentence \textit{S = $w_1$, $w_2$, ...,$w_{|S|}$,} then we formulate a probability distribution over $S$ in such a way that at each time step $t_k$ during the training phase, we train the model to maximize the posterior probability of sub-sequence \textit{$w_1$, $w_2$,... $w_{k-1}$}, given the image $I$.  The probability distribution can be stated as:\\
\begin{large}
\begin{equation} \label{eq1}
    P(w_1,w_2,...,w_{k} \arrowvert I)= \prod_{t=1}^{k} P(w_t  \arrowvert I,w_{0:t-1})
\end{equation}
\end{large} 
with $w_0$ being the $<START>$ token which is pre-fixed with the sequence as an signal to begin the generation process. Hence, the generation of next word is conditioned on the words \begin{large} $w_{0:t-1}$ \end{large} that have been generated previously and the input image. Since the Equation \ref{eq1} is recursive in nature, the solution can be implemented as a Recurrent Neural Network decoder. Each factor in the equation represents one step of the word generation process.

The \textbf{core contributions} of this work are as follows:
\begin{itemize}
    \item We utilise Inception-ResNet-v2 (Szegedy et. al.\cite{incepres}) as encoder for extracting the image feature representation, and a Deep LSTM (Hochreiter et. al.\cite{lstm}) network for generating the caption. Compared with other similar approaches, feature extraction using Inception-ResNet-v2 allows us to gain significant improvements in performance of our models. Also the Deep LSTM network allows learning of more complex relationships as compared to a single layered network. 
    \item We employ pre-trained word embeddings proposed in Komninos et. al. \cite{komninos} which employ word and dependency context features to generate the word representations. To do this, the authors consider word co-occurrences between  pairs  of  words,  words  and  dependency  context features, and between different dependency context features. We explain the word embeddings in detail in Section \ref{embeddings}.
    \item In order to further improve the performance of system, we fine-tune the parameters of Word Embedding layer and the Feature Extractor (CNN) layer. We perform fine-tuning after our models shows breakdown of co-relation between validation set log-likelihood error and the image captioning evaluation metrics (BLEU-4 used in our experiments):
    \begin{itemize}
        \item We fine-tune the words embeddings (that we use as word representations). In other words, unlike the previous works we do not use the pre-trained word vectors as fixed weights in the word embedding layer to generate a feature representation for input words. Instead, we allow the network to learn the weights of the word embedding layer after initializing the weights with the pre-trained word embeddings proposed in Komninos et. al.\cite{komninos}. We adopt a learning rate for the word embedding layer which is 10 times lower than the learning rate of the decoder layer. 
        \item Since we are using transfer learning with pre-trained CNN to extract features from the images,  following the observations of Yosinski et. al. \cite{yosinski} where they find that transferability of Deep Neural Networks is negatively affected primarily by the specialization of higher layer neurons, we perform fine-tuning on the CNN. We freeze the weights of the first few layers and allow the model to learn the weights of higher layers using back-propagation. The learning rate for the CNN is 10 times lower than the learning rate for the decoder layer.
    \end{itemize}
    
    \item We use image data augmentation to prevent overfitting since the datasets that are commonly used in Image Captioning have very limited number of images and their annotated descriptions. Image Data Augmentation has been used in Image Captioning in Wang et. al. \cite{bilstm} where the authors have explored random cropping of the images, multi-scaling and vertical flipping of the images. However, cropping the images may leave out some important information from some images where the objects are present close to the edges of the images. Also, vertical flipping the images may not account for spatial relationships among the objects between the training data and test set as data in the test set will not be augmented. Hence, we explore horizontal flipping of the images and Perspective Transformation to create more image-sentence pairs. Horizontal flipping preserves more spatial relationships and perspective transformation of images introduces slight distortion in images with respect to the reference plane of the image but preserves all spatial relationships. We explain Image Data Augmentation in detail in Section \ref{dataaug}.
\end{itemize}

We evaluate our approach with two generation methods:\\ 
(a) Encoder-Decoder Approach or Inject Approach: The image feature representations are used as initial hidden and cell states of the decoder and decoder generates the caption, one word a time, by taking the generated caption sub-sequence until the last time step. This approach is similar to the approach proposed in Vinyals et. al.\cite{vinyals}. However, our approach differs from their approach in following respects:
\begin{itemize}
    \item We extract a set of feature vectors from an intermediate convolutional layer of the CNN as explained in Section \ref{incepres} unlike extracting a feature vector from the final fully connected layer of the CNN as has been done in Vinyals et. al.\cite{vinyals}. We have found that this approach aids in generating better captions because it contains information about the spatial relationships between objects in the image.  
    \item As mentioned earlier, we utilise Inception-ResNet-v2 CNN as encoder and Deep LSTM network as decoder along with word embeddings proposed in Komninos et. al. \cite{komninos} unlike the VGG-16 encoder and a single LSTM layered decoder in Vinyals et. al.\cite{vinyals}. Also the authors of Vinyals et. al. \cite{vinyals} don't use pre-trained word embeddings as they find that using pretrained embeddings from Mikolov et. al. \cite{mikolov} doesn't improve the performance of their model.
    \item The work of Vinyals et. al. \cite{vinyals} uses a model ensemble which allows them to achieve a performance gain of a few points worth in terms of BLEU score. Since we evaluate our approach with respect to two methods (Encoder-Decoder Approach and Soft Attention), in order to ensure a fair comparison between the two, we do not use model ensembles.
    \item And lastly, unlike Vinyals et. al. \cite{vinyals}, we do not use batch normalization on the inputs because it does not help us to achieve tangible performance gains of our models. One reason for this may be that, since batch normalization is a technique to prevent overfitting of the model to training data, use of data augmentation helps us to alleviate overfitting and thus batch normalization is no longer needed.
\end{itemize}
(b) Attention based approach: The image feature representations, after being mapped to the same vector space as the hidden representations of the LSTM layer, are used as first state of the decoder. This is similar to the Encoder-Decoder approach, but at each time step the decoder also gets, as input, a context vector, $\widetilde{\textbf{s}}_{t}$ which is a dynamic representation of the input image at that time step containing information about the relevant portions of the image. Our attention-based method is similar to the method proposed by Xu et. al. \cite{attendtell} with some differences:
\begin{itemize}
    \item We utilise Inception-ResNet-v2 as encoder and Deep LSTM network as decoder along with word embeddings proposed in Komninos et. al. \cite{komninos} unlike the VGG-16 encoder and a single LSTM layered decoder used by Xu et. al.\cite{attendtell}. In addition, for feeding the text data into the Deep LSTM layer, we use the word representations proposed in Komninos et. al\cite{komninos}. 
\end{itemize}

\section{Related Work}
\label{related}
\subsection{Image Caption Generation}
\label{captioning}
We can classify the approaches used for image caption generation into three categories. In first category are the template-based approaches which use caption templates and generate the captions according to a template after detecting objects and recognizing attributes in an input image. As an example, Li et. al. \cite{li} first learn the relationships between different objects present in an image and then parse a sentence into multiple phrases followed by learning relationships between objects in the image and constituent phrases in the sentence. Kulkarni et. al. \cite{kulkarni}, use conditional random field (CRF) to correspond to the objects, attributes, and prepositions in the image and try to learn the most appropriate label. Template based methods have to be hand-designed and have rigid templates. This limits the number of possible sentences that can be generated thereby leading to poor performance. Also generating sentences with variable-length is very difficult with this approach.

Retrieval-based approaches fall into the second category. These methods approach image captioning as a retrieval task where they retrieve images with similar captions and then modify and combine the captions to produce new captions. A distance metric is used to find images that have similar captions, such as the method proposed in Kuznetsova et. al. \cite{kuznetsova}. In addition, these approaches also have an intermediate “generalization” step to fine-tune the caption by discarding the information within the caption that was a carryover from the retrieved images and is not relevant to the image for which the caption is being generated, such as the name of a landmark or a type of food. 

Recently, Deep Neural Network based approaches have been adopted which have provided state of the art results over the last few years. These are mostly due to success of CNNs in object recognition task, as in Krizhevsky et. al. \cite{krizhevsky}, and the success of Recurrent Neural Networks (RNNs) on the sequence to sequence tasks, particularly with the advent of machine translation with RNNs as in Sutskever et. al. \cite{sutskever} and Bahdanau et. al.\cite{bahdanau}. Sutskever et. al.\cite{sutskever} proposed the encoder-decoder format of machine translation using an RNN for encoding the text information into a context vector and another RNN which decoded the context vector and produces the caption recursively. Whereas Bahdanau et. al. \cite{bahdanau} used attention mechanism to focus on certain regions of the input sentence so that decoder has more relevant information at a certain time-step. Both these models are the basis of work done in Image Captioning models proposed by Vinyals et. al. \cite{vinyals} and Xu et. al. \cite{attendtell}, respectively. For Image Captioning, the pioneering work using the encoder-decoder framework was done by Kiros et. al. \cite{kiros} who used Structure Content Neural Language Model (SC-NLM) as the decoder. Socher et. al. \cite{socher} proposed a Dependency Tree Recursive Neural Network (DT-RNN) which was used for embedding the sentence into a vector space which is then used to retrieve images. Mao et. al. \cite{mao} proposed the Multimodal Recurrent Neural Network (m-RNN) where the word embeddings and image feature representations are mapped into the same vector space and merged before being used as input to an RNN which then decodes the sentence one step at a time. Vinyals et. al. \cite{vinyals} proposed a system in which word embeddings are mapped into the same space as the hidden representations of the LSTM and used as the initial hidden and cell states of the LSTM. In subsequent steps the input is provided in the form of embedding vectors of the previously generated words and the next word in the sequence is generated. Subsequently, Xu et. al. \cite{attendtell} used attention mechanism to focus on the salient regions of the image. In You et. al.\cite{you}, semantic attention model is used where top-down and bottom-up approaches are combined for generating captions with Recurrent Neural Networks. While using the bottom-up approach, semantic concepts are used as attributes whereas in the top-down approach, the authors employ visual features with attention mechanism to guide the RNN. More recently, following the work of Gehring et. al. \cite{gehring} for generating sequences with Convolutional Neural Networks, Aneja et. al. \cite{aneja} have proposed using two Convolutional Neural Networks where the first network encodes the image information and the next one acts as a decoder to generate the caption sentence.
\subsection{Word Embeddings}
\label{embeddings}
The simplest approach to process text in NLP systems is to treat words as atomic units where they are represented as indices in a vocabulary. In this approach there is no focus on the similarity between words. This has some advantages of simplicity and robustness. Moreover, it has been observed that simple models trained on huge amounts of data perform better than complex systems trained on less data. For example, the N-gram models which are employed in statistical language processing are trained N-grams on virtually huge amounts of data in the range of trillions of words (Brants et. al. \cite{brants}).
However, these simple techniques are not always viable. In a lot of cases, the amount of data available isn't enough to train such simple models. Even in machine translation, the entire literature available for many low-resource languages contain only a few billions of words or less and the annotated data for translation or captioning tasks contains even fewer words. For example, the Flickr8k dataset \cite{flickr8k} used for Image Captioning has around 40000 sentences in total with most sentences having lengths of less than 20 words (with 34 being the largest sentence length), thus bringing the total number of words to be far lesser than 1 million words. Hence, the techniques which learn simple relations between words, such as word frequency statistics, will not result in any significant progress.
Hence, its imperative that methods which attempt to learn complex relationships between words are used. Probably the first concept which employed distributed representations of words is provided in Hinton et. al.\cite{hinton}. However, with the progress of machine learning the widespread use has started only in recent years. The pioneering work in this regard is by Mikolov et. al.\cite{mikolov} where the authors propose to represent words as dense vectors (as opposed to sparse one hot vectors).
They propose two variants: Continuous bags of Words (CBOW) and Skip-Gram Model. In the CBOW model, the inputs are the merged representations of words that appear in the surrounding context of a particular word and the desired output is the word whose context is being explored. In the Skip Gram model the Neural network tries to predict the words surrounding a given word(which is the input).

Word embeddings are a better approach towards generalization to unseen examples because they are able to incorporate syntactic and semantic properties of words. For the task of Image Captioning, word embeddings have been used as a separate embedding layer which takes as input a sequence of words which have been encoded as numbers corresponding to the position of words in the vocabulary and produces as output a vector of length $n$ where $n$ is the embedding length. The weights of this embedding layer are either learned through backpropagation or are adapted from other language modelling tasks. Pretrained word embeddings have been used as well, such as in You et. al. \cite{you} who use pretrained Glove word vectors from Pennington et. al. \cite{glove} to encode word information into the LSTM. Glove embeddings are an improvement to the skip-gram model proposed in Mikolov et. al.\cite{mikolov} but both use the linear sequential properties of the text. In other words, both use the word co-occurrence properties of the text to calculate representations of the text. Other works such as Vinyals et. al. \cite{vinyals} have noted that using pretrained embeddings does not provide performance enhancement in their model.

We propose to employ word embeddings proposed in Komninos et. al.\cite{komninos} which employ both word and dependency context features. They consider co-occurrences between pairs of words, words and dependency context features, and between different dependency context features in a dependency graph . This is an extension to the dependency graph based embeddings proposed by Levy et. al.\cite{levy}. Hence the representations obtained share the properties common to both the window based models and the models based on dependency graphs. In addition, this approach provides additional structural information for the dependency graph based embeddings.
Similar to window based models where the words appearing within a certain distance are considered, in dependency based embeddings, a target word is considered as a node in the dependency graph. Then during the training process word embeddings are optimized in such a way that they maximize the probability of other words occurring within a certain distance in the graph.

\subsection{Data Augmentation}
\label{aug}
Data Augmentation, as discussed in Taylor et. al.\cite{taylor}, encompasses a set of techniques which aim to increase the size and improve the quality of datasets that are used for Deep Learning techniques. The image data augmentation techniques aim to prevent over-fitting of the model to the training set in such a way that the model effectively 'sees' more data. Two possible approaches to this end are data warping and oversampling. Data warping involves transforming the existing images in the dataset in such a way that they appear to be sufficiently different that the existing ones but contain similar content so that their label is preserved. This includes techniques like geometric transformations such as flipping or rotating the images, shearing or scaling them or perspective transformations, color transformations such as changing the contrast, brightness or color of objects, cropping or random erasing of images,neural style transfer and adversarial training. Oversampling techniques for data augmentation produce synthetic instances of the data and add them to the training set. The techniques include mixing the images, generative adversarial networks (GANs) and augmentations to the feature space. In Image Captioning, Data Augmentation has been used in the previous work in Wang et. al.\cite{bilstm} where the authors use Multi-Crop (cropping from four corners and center region), Multi-Scale (where the image is first resized to a fixed size, and then the authors select a region of the image with height and width calculated by multiplying the original dimensions by a randomly chosen constant from a fixed set of constants). 

In our work, we adopt Random Horizontal Flipping, Random Vertical Flipping and Random Perspective Transformation as described in detail in Section \ref{dataaug}. Random Perspective Transforms have been used in previous works in some image processing tasks. For example, it has been used by Lin et. al.\cite{lin} in estimation of the number of persons in a crowded environment. We have found that Perspective Transforms coupled with horizontal and vertical flipping help us improve the performance of our system by a few points in terms of evaluation metrics.

\section{Proposed Method}
\label{proposed}
\begin{figure}
\tikzstyle{startstop} = [rectangle, rounded corners, minimum width=1cm, minimum height=1cm,text centered, draw=black]
\tikzstyle{io} = [trapezium, trapezium left angle=70, trapezium right angle=110, minimum width=1cm, minimum height=.5cm, text centered, draw=black]
\tikzstyle{input} = [rectangle, minimum width=1.5cm, minimum height=0.5cm, text centered, draw=black]
\tikzstyle{LSTM} = [rectangle, rounded corners, minimum width=2cm, minimum height=2cm, text centered, draw=black, fill=blue!20]
\tikzstyle{SOFT} = [rectangle, rounded corners, minimum width=2cm, minimum height=1cm, text centered, draw=black, fill=blue!10]
\tikzstyle{decision} = [diamond, minimum width=1cm, minimum height=.2cm, text centered, draw=black]
\tikzstyle{fcimagestyle} = [rectangle, rounded corners, minimum width=1cm, minimum height=1cm,text centered, draw=black, fill=red!30]
\tikzstyle{fctextstyle} = [rectangle, rounded corners, minimum width=1cm, minimum height=1cm,text centered, draw=black, fill=orange!20]
\tikzstyle{embedstyle} = [rectangle, rounded corners, minimum width=1cm, minimum height=1cm,text centered, draw=black, fill=orange!40]
\tikzstyle{NASNET} = [rectangle, rounded corners, minimum width=3cm, minimum height=1.6cm,text centered, draw=black, fill=olive!30]
\tikzstyle{dotbox} = [rectangle, rounded corners, minimum width=1cm, minimum height=1cm,text centered, draw=none]
\tikzstyle{atimagestyle} = [rectangle, rounded corners, minimum width=1cm, minimum height=1cm,text centered, draw=black, fill=brown!30]

\tikzstyle{arrow} = [thick,->]
\centering
\resizebox{1.0\textwidth}{!}{%
\begin{tikzpicture}[node distance = 2.0cm]
    \node (lstm1) [LSTM]
        {\Huge{LSTM (i)  $t_{1}$}} ;
    \node (lstm1a)[LSTM, above of=lstm1, node distance=3cm] {\Huge{LSTM (ii) $t_{1}$ }};
    \node (lstm1b)[LSTM, above of=lstm1a, node distance=3cm] {\Huge{LSTM (iii) $t_{1}$ }};
    \node (emb0) [embedstyle, below of=lstm1, node distance=3cm] {\begin{tabular}{c}
        \Huge{E L $t_{1}$} \\
    \end{tabular} };
    \node (start0) [input, below of=emb0]{\huge{$<$START$>$}};
    \node (soft1) [SOFT, above of=lstm1b, node distance=2.6cm]{\huge{SOFTMAX}};
    \node (wordone0) [input, above of=soft1, node distance=2.0cm]{\Huge{$word_1$}};
    \node (fcimage) [fcimagestyle, left of=lstm1, node distance=4.5cm,  xshift=0cm] {\begin{tabular}{c}
        \Huge{F C}\\
    \end{tabular}};

    \node (nasnet)[NASNET, below of=fcimage, node distance=2.5cm] {\begin{tabular}{c}
         \Huge{Encoder}
    \end{tabular}};
    \node (image)[input, below of=nasnet, node distance=3.0cm]{\begin{tabular}{c}
         \Huge{Input}  \\
         \Huge{Image} 
    \end{tabular}};
    
    \node (serial) [dotbox, below of=image, xshift=2.0cm, yshift=0cm] {\Huge{(a) Encoder-Decoder Approach}};
    \node (ell) [embedstyle, below of=serial, xshift=0.0cm, yshift=0cm] {\Huge{ EL}};
    \node (ella) [dotbox, right of=ell, xshift=2.50cm, yshift=0cm] {\Huge{Embedding Layer}};
    \node (fcl) [fcimagestyle, below of=ell, xshift=0.0cm, yshift=0cm] {\Huge{ FC}};
    \node (fcla) [dotbox, right of=fcl, xshift=2.30cm, yshift=0cm] {\Huge{Fully Connected}};
    \node (all) [atimagestyle, below of=fcl, xshift=0.0cm, yshift=0cm] {\Huge{ AL}};
    \node (alla) [dotbox, right of=all, xshift=2.30cm, yshift=0cm] {\Huge{Attention Layer}};

    \node (lstm2) [LSTM, right of=lstm1, node distance=6cm] {\begin{tabular}{c}
        \Huge{LSTM (i) $t_{2}$}
    \end{tabular} };
    \node (lstm2a)[LSTM, above of=lstm2, node distance=3cm] {\Huge{LSTM (ii) $t_{2}$ }};
    \node (lstm2b)[LSTM, above of=lstm2a, node distance=3cm] {\Huge{LSTM (iii) $t_{2}$ }};
    \node (emb1) [embedstyle, below of=lstm2, node distance=3cm] {\begin{tabular}{c}
        \Huge{E L $t_{2}$} \\
    \end{tabular} };
    \node (start) [input, below of=emb1]{\Huge{$word_{1}$}};
    \node (soft2) [SOFT, above of=lstm2b, node distance=2.6cm]{\huge{SOFTMAX}};
    \node (wordone) [input, above of=soft2, node distance=2.0cm]{\Huge{$word_2$}};

    \node (dots) [dotbox, right of=lstm2, node distance=3.5cm]{\Huge{\textbf{. . .}}};
    \node (dots) [dotbox, right of=lstm2a, node distance=3.5cm]{\Huge{\textbf{. . .}}};
    \node (dots) [dotbox, right of=lstm2b, node distance=3.5cm]{\Huge{\textbf{. . .}}};

    \node (lstmn) [LSTM, right of=lstm2, node distance=7cm] {\begin{tabular}{c}
        \Huge{LSTM (i) $t_{k}$}
    \end{tabular} };
    \node (lstmna)[LSTM, above of=lstmn, node distance=3cm] {\Huge{LSTM (ii) $t_{k}$ }};
    \node (lstmnb)[LSTM, above of=lstmna, node distance=3cm] {\Huge{LSTM (iii) $t_{k}$ }};

    \node (embn) [embedstyle, below of=lstmn, node distance=3cm] {\begin{tabular}{c}
        \Huge{E L $t_{k}$} \\
    \end{tabular} };
    \node (wordlastinp) [input, below of=embn]{\Huge{$word_{k-1}$}};
    \node (soft) [SOFT, above of=lstmnb, node distance=2.6cm]{\huge{SOFTMAX}};
    \node (wordn) [input, above of=soft, node distance=2cm]{\Huge{$word_{k}$}};

    \draw [arrow, ->] (lstm1) |- (lstm2);
    \draw [arrow, ->] (lstm1a) |- (lstm2a);
    \draw [arrow, ->] (lstm1b) |- (lstm2b);
    \draw [arrow, ->] (fcimage) -- (lstm1);
    \draw [arrow, ->] (fcimage) -- (-4.65,6.0);
    \draw [arrow, ->] (-4.65,3.0) -- (lstm1a);
    \draw [arrow, ->] (-4.65,6.0) -- (lstm1b);
    
    \draw[arrow] (nasnet) -- (fcimage);
    \draw [arrow] (image) -- (nasnet);
    \draw [arrow] (start0) -- (emb0);
    \draw [arrow] (emb0) -- (lstm1);
    \draw [arrow] (lstm1) -- (lstm1a);
    \draw [arrow] (lstm1a) -- (lstm1b);
    \draw [arrow] (lstm1b) -- (soft1);
    \draw [arrow] (soft1) -- (wordone0);

    \draw [arrow, ->] (emb1) -- (lstm2); 
    \draw [arrow, ->] (start) -- (emb1);
    \draw [arrow] (lstm2) -- (lstm2a);
    \draw [arrow] (lstm2a) -- (lstm2b);
    
    \draw [arrow] (lstm2b.north) -- (soft2.south);
    \draw [arrow] (soft2.north) -- (wordone.south);

    \draw [arrow, ->] (embn) -- (lstmn); 
    \draw [arrow, ->] (wordlastinp) -- (embn); 
    \draw [arrow] (lstmn) -- (lstmna);
    \draw [arrow] (lstmna) -- (lstmnb);
    \draw [arrow, ->] (lstmnb.north) to (soft.south);
    \draw [arrow] (soft)--(wordn);

    \begin{scope}[xshift=23cm, yshift=0cm]
    \node (lstm1) [LSTM]
        {\Huge{LSTM (i)  $t_{1}$}} ;
    \node (lstm1a) [LSTM, above of=lstm1, node distance=3cm] {\Huge{LSTM (ii)  $t_{1}$}};
    \node (lstm1b) [LSTM, above of=lstm1a, node distance=3cm] {\Huge{LSTM (iii) $t_{1}$}} ;
     \node (sum1) [dotbox, below of=lstm1, node distance = 2.5cm]{\huge{$\bigoplus$}};
    \node (emb0) [embedstyle, below of=lstm1, node distance=6cm, xshift=1cm] {\begin{tabular}{c}
        \Huge{E L $t_{1}$} \\
    \end{tabular} };
    \node (at1) [atimagestyle, below of=lstm1, node distance= 4.2cm, xshift=-1.8cm]{\huge{AL $t_{1}$}};

    \node (start0) [input, below of=emb0]{\huge{$<$START$>$}};
    \node (soft1) [SOFT, above of=lstm1b, node distance=2.6cm]{\huge{SOFTMAX}};
    \node (wordone0) [input, above of=soft1, node distance=2.0cm]{\Huge{$word_1$}};
    
    \node (fcimage) [fcimagestyle, left of=lstm1, node distance=5.0cm,  xshift=0cm] {\begin{tabular}{c}
        \Huge{F C}\\
    \end{tabular}};
    
    \node (nasnet)[NASNET, below of=fcimage, node distance=9.5cm] {\begin{tabular}{c}
         \Huge{Encoder}
    \end{tabular}};
    \node (image)[input, below of=nasnet, node distance=3.0cm]{\begin{tabular}{c}
         \Huge{Input}  \\
         \Huge{Image} 
    \end{tabular}};
    \node (arrowdesc1) [dotbox, right of=lstm1, xshift=0.3cm, yshift=0.5cm] {};
    \node (serial) [dotbox, below of=image, xshift=2.0cm, yshift=0cm] {\Huge{(b) Soft Attention}};
    
    \node (lstm2) [LSTM, right of=lstm1, node distance=6cm] {\begin{tabular}{c}
        \Huge{LSTM (i) $t_{2}$}
    \end{tabular} };
    \node (lstm2a) [LSTM, above of=lstm2, node distance=3cm] {\Huge{LSTM (ii)  $t_{2}$}};
    \node (lstm2b) [LSTM, above of=lstm2a, node distance=3cm] {\Huge{LSTM (iii) $t_{2}$}} ;
    \node (sum2) [dotbox, below of=lstm2, node distance = 2.5cm]{\huge{$\bigoplus$}};
    \node (emb1) [embedstyle, below of=lstm2, node distance=6cm,xshift=1cm] {\begin{tabular}{c}
        \Huge{E L $t_{2}$} \\
    \end{tabular} };
    \node (start) [input, below of=emb1]{\Huge{$word_{1}$}};
    \node (soft2) [SOFT, above of=lstm2b, node distance=2.6cm]{\huge{SOFTMAX}};
    \node (wordone) [input, above of=soft2, node distance=2.0cm]{\Huge{$word_2$}};
    
    \node (at2) [atimagestyle, below of=lstm2, node distance= 4.2cm, xshift=-1.8cm]{\huge{AL $t_{2}$}};

    \node (dots) [dotbox, right of=lstm2, node distance=3.5cm]{\Huge{\textbf{. . .}}};
    \node (dots) [dotbox, right of=lstm2a, node distance=3.5cm]{\Huge{\textbf{. . .}}};
    \node (dots) [dotbox, right of=lstm2b, node distance=3.5cm]{\Huge{\textbf{. . .}}};

    \node (lstmn) [LSTM, right of=lstm2, node distance=7cm] {\begin{tabular}{c}
        \Huge{LSTM (i) $t_{k}$}
    \end{tabular} };
    \node (lstmna) [LSTM, above of=lstmn, node distance=3cm] {\Huge{LSTM (ii)  $t_{k}$}};
    \node (lstmnb) [LSTM, above of=lstmna, node distance=3cm] {\Huge{LSTM (iii) $t_{k}$}} ;
    \node (sumn) [dotbox, below of=lstmn, node distance = 2.5cm]{\huge{$\bigoplus$}};
    \node (embn) [embedstyle, below of=lstmn, node distance=6cm,xshift=1cm] {\begin{tabular}{c}
        \Huge{E L $t_{k}$} \\
    \end{tabular} };
    \node (wordlastinp) [input, below of=embn]{\Huge{$word_{k-1}$}};
    \node (soft) [SOFT, above of=lstmnb, node distance=2.6cm]{\huge{SOFTMAX}};
    \node (wordn) [input, above of=soft, node distance=2cm]{\Huge{$word_{k}$}};
    \node (atn) [atimagestyle, below of=lstmn, node distance= 4.2cm, xshift=-1.8cm]{\huge{AL $t_{k}$}};

    \draw [arrow, ->] (lstm1) |- (lstm2);
    \draw [arrow, ->] (lstm1a) |- (lstm2a);
    \draw [arrow, ->] (lstm1b) |- (lstm2b);
    
    \draw [arrow, ->] (fcimage) -- (lstm1);
    \draw [arrow, ->] (fcimage.north) -- (-5,6.0);
    \draw [arrow, ->] (-5.0,3.0) -- (lstm1a);
    \draw [arrow, ->] (-5.0,6.0) -- (lstm1b);
    
    \draw[arrow] (nasnet) -- (fcimage);
    \draw [arrow] (image) -- (nasnet);
    \draw [arrow] (start0) -- (emb0);
    
    \draw [arrow, ->] (emb0.north) |- (sum1.east);
    \draw [arrow, ->] (sum1) |- (lstm1.south);
    
    \draw [arrow, ->] (lstm1.north) to (lstm1a.south);
    \draw [arrow, ->] (lstm1a.north) to (lstm1b.south);
    
    \draw [arrow] (lstm1b.north) -- (soft1.south);
    \draw [arrow] (soft1.north) -- (wordone0.south);
    
    \draw [arrow] (nasnet.east) -| (11.2,-9.5);
    \draw [arrow] (11.2,-9.5) -- (atn);
    \draw [arrow] (4.2,-9.5) -- (at2);
    \draw [arrow] (-1.8,-9.5) -- (at1);
    
    \draw [arrow] (at1.north) |- (sum1.west);
    \draw [arrow] (at2.north) |- (sum2.west);
    \draw [arrow] (atn.north) |- (sumn.west);

    \draw [arrow, ->] (emb1.north) |- (sum2.east);
    \draw [arrow, ->] (start) -- (emb1); 
    \draw [arrow, ->] (sum2) |- (lstm2.south);
    \draw [arrow, ->] (lstm2.north) to (lstm2a.south);
    \draw [arrow, ->] (lstm2a.north) to (lstm2b.south);
    
    \draw [arrow] (lstm2b.north) -- (soft2.south);
    \draw [arrow] (soft2.north) -- (wordone.south);

    \draw [arrow, ->] (embn.north) |- (sumn.east);
    \draw [arrow, ->] (sumn) |- (lstmn.south);
    \draw [arrow, ->] (wordlastinp) -- (embn); 
    \draw [arrow, ->] (lstmn.north) to (lstmna.south);
    \draw [arrow, ->] (lstmna.north) to (lstmnb.south);
    \draw [arrow, ->] (lstmnb.north) to (soft.south);
    \draw [arrow] (soft)--(wordn);
    \end{scope}
    
\end{tikzpicture}
}
\captionsetup{format=default,justification=justified}
\centering
\caption{An overview of the Image Caption Generation process used in our work. The process depicted here is for the time-step $k$. If we need to generate the output at k-th timestep, we feed all words up to $k-1$ timesteps. The output of fully connected layers $f_{init\_c}$ and $f_{init\_h}$, as explained in Section \ref{incepres} (and depicted here with a single Fully Connected Layer), is fed into the LSTM at the initial time-step. Both the Encoder-Decoder Approach and Soft Attention frameworks are shown here in Figure (a) and (b) respectively.
Here, $\bigoplus$ stands for concatenation operation.
}
\label{overview}
\end{figure}
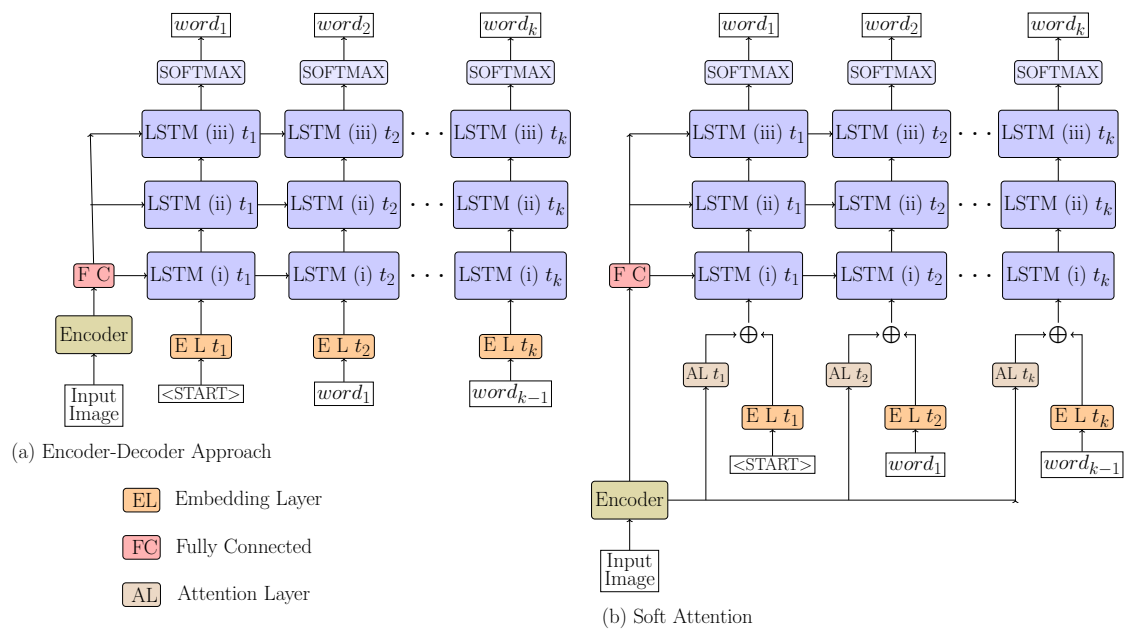

In this section we first describe the common components of the two methods that we evaluate and then discuss the individual details. In both the Encoder-Decoder and Soft Attention method, we use InceptionResNet CNN to extract a set of feature representations from a lower Convolutional layer of the CNN. And we use a Deep Multi Layer LSTM network to generate the captions one word at a time.
For each input image, both the \textit{Encoder-Decoder} and \textit{Soft Attention} approaches produce an output caption $y$ which is a sequence of words encoded as \textit{1-of-K} vectors,
\begin{equation}
    y = \{y_1, y_2, y_3, . . . , y_n \}, y_i \in R^K
\end{equation}
where \textit{n} is the maximum length of the captions and \textit{K} is the length of the vocabulary.
We explain both the approaches in Figure \ref{overview}.

\subsection{Embedding Layer}
\label{embedding}
We use pre-trained Word Embeddings from Komninos et. al.\cite{komninos} for the input sequence of words. For each word, we get an embedding vector of length $m$ where $m=300$ in this case. Hence for a sequence of $n$ words we get a  \begin{math} n \times m \end{math} matrix as an input to the LSTM where $n$ is the length of the sequence and $m$ is the number of features in the input. More formally, we can state that
\begin{equation}
X = {x_1,x_2, ..., x_n}    
\end{equation}
where $n$ is the length of the sequence, the embedding layer generates a sequence
\begin{equation}\label{eqp}
    E = {e_1, e_2, ..., e_n}
\end{equation}

where $n$ is the length of the sequence and  \begin{math}e_i \in R^{m \times K} \end{math} where $m$ is the embedding dimension and $K$ is the size of the vocabulary.

\subsection{InceptionResNet CNN:}
\label{incepres}
We use Inception-ResNet Convolutional Neural Network \cite{incepres} for extracting features from input features. Unlike the VGG Net used in the works of Vinyals et. al. \cite{vinyals} and Xu et. al. \cite{attendtell}, the Inception-ResNet architecture performs better on the ImageNet object detection challenge \cite{imagenet}. Using Inception ResNet, we extract features from a lower convolutional layer which produces $V$ vectors of $D$ dimensions each such that each of them is a representation of some part of the image. Stated formally, we extract an image representation $p$ such that,
\begin{equation}
    p = \{p_1,p_2,p_3, . . . ,p_V\}, p_i \in R^D 
\end{equation}
Since it is a set of $V$ vectors, it allows the decoder to focus on different parts of the image.
This set of vectors is used to calculate the set of initial hidden and cell states for the LSTM decoder (explained in Section \ref{lstm_section}) and in the Soft Attention model (Section \ref{softatten}) to guide the decoder to focus on certain regions of the image.

\subsection{Long Short Term Memory Network}
\label{lstm_section}
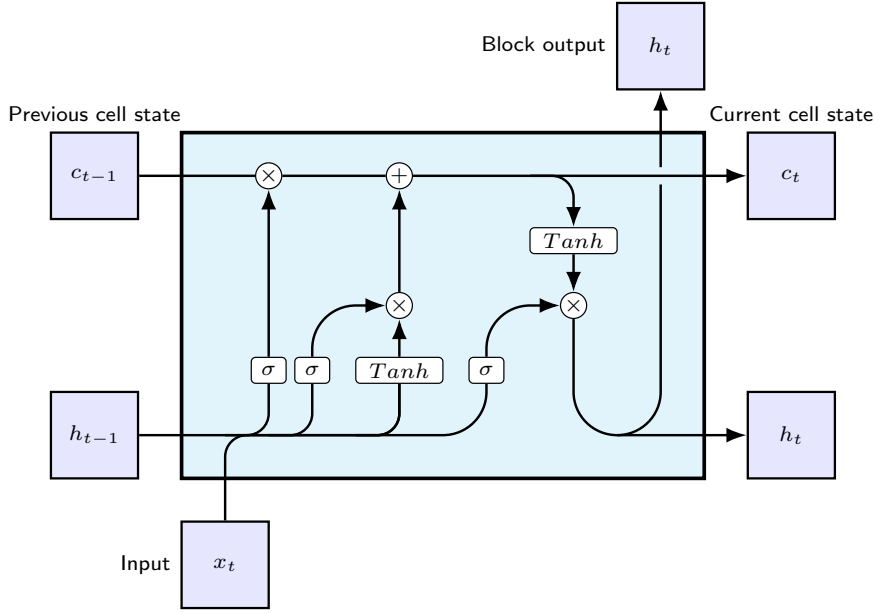
\begin{figure}[h]
\centering
\resizebox{0.8\textwidth}{!}{
\begin{tikzpicture}[
    font=\sf \scriptsize,
    >=LaTeX,
    cell/.style={
        rectangle, 
        sharp corners=5mm, 
        draw,
        very thick,
        fill=cyan!10,
        },
    operator/.style={
        circle,
        draw,
        inner sep=-0.5pt,
        minimum height =.3cm,
        fill=white!100,
        },
    function/.style={
        ellipse,
        draw,
        inner sep=1pt
        },
    ct/.style={
        rectangle,
        draw,
        line width = .75pt,
        minimum width=1cm,
        minimum height=1cm,
        inner sep=1pt,
        fill=blue!10},
    gt/.style={
        rectangle,
        rounded corners=0.5mm,
        draw,
        minimum width=4mm,
        minimum height=3mm,
        inner sep=1pt,
        fill=white!15},
    mylabel/.style={
        font=\scriptsize\sffamily
        },
    ArrowC1/.style={
        rounded corners=.25cm,
        thick,
        },
    ArrowC2/.style={
        rounded corners=.5cm,
        thick,
        },
    ]

    \node [cell, minimum height =4cm, minimum width=6cm] at (0,0){} ;

    \node [gt] (ibox1) at (-2,-0.75) {$\sigma$};
    \node [gt] (ibox2) at (-1.5,-0.75) {$\sigma$};
    \node [gt, minimum width=1cm] (ibox3) at (-0.5,-0.75) {$Tanh$};
    \node [gt] (ibox4) at (0.5,-0.75) {$\sigma$};

    \node [operator] (mux1) at (-2,1.5) {$\times$};
    \node [operator] (add1) at (-0.5,1.5) {+};
    \node [operator] (mux2) at (-0.5,0) {$\times$};
    \node [operator] (mux3) at (1.5,0) {$\times$};
    \node [gt, minimum width=1cm] (func1) at (1.5,0.75) {$Tanh$};

    \node[ct, label={[mylabel]Previous cell state}] (c) at (-4,1.5) {$c_{t-1}$};
    \node[ct, label={[mylabel]}] (h) at (-4,-1.5) {$h_{t-1}$};
    \node[ct, label={[mylabel]left:Input}] (x) at (-2.5,-3) {$x_{t}$};

    \node[ct, label={[mylabel]Current cell state}] (c2) at (4,1.5) {$c_t$};
    \node[ct, label={[mylabel]}] (h2) at (4,-1.5) {$h_t$};
    \node[ct, label={[mylabel]left:Block output}] (x2) at (2.5,3) {$h_t$};

    \draw [ArrowC1] (c) -- (mux1) -- (add1);
    \draw [->,ArrowC2] (add1) -- (c2);
    
    \draw [ArrowC2] (h) -| (ibox4);
    \draw [ArrowC1] (h -| ibox1)++(-0.5,0) -| (ibox1); 
    \draw [ArrowC1] (h -| ibox2)++(-0.5,0) -| (ibox2);
    \draw [ArrowC1] (h -| ibox3)++(-0.5,0) -| (ibox3);
    \draw [ArrowC1] (x) -- (x |- h)-| (ibox3);
    
    \draw [->, ArrowC2] (ibox1) -- (mux1);
    \draw [->, ArrowC2] (ibox2) |- (mux2);
    \draw [->, ArrowC2] (ibox3) -- (mux2);
    \draw [->, ArrowC2] (ibox4) |- (mux3);
    \draw [->, ArrowC2] (mux2) -- (add1);
    \draw [->, ArrowC1] (add1 -| func1)++(-0.5,0) -| (func1);
    \draw [->, ArrowC2] (func1) -- (mux3);

    \draw [->, ArrowC2] (mux3) |- (h2);
    \draw (c2 -| x2) ++(0,-0.1) coordinate (i1);
    \draw [-, ArrowC2] (h2 -| x2)++(-0.5,0) -| (i1);
    \draw [->, ArrowC2] (i1)++(0,0.2) -- (x2);

\end{tikzpicture}}
\caption{Illustration of a basic LSTM cell.}
\label{lstm}
\end{figure}

We use a Deep Long Short Term Memory (LSTM) network composed of LSTM layers (Hochreiter et. al.\cite{lstm}) stacked on top of each other such that output of the previous LSTM layer is used as input for the next layer. We provide the basic description of an LSTM in Figure \ref{lstm}. In an LSTM cell there is a single cell state and three gates which the input, output, and forget gates. During each time-step $t$, the generated cell state $c_{t-1}$ and the hidden state $h_{t-1}$ which were generated during previous time-step $t-1$ , are forwarded back to the LSTM. The input $u_t$ is received at present time $t$. If we use $f_{LSTM}$(·) to represent feed-forward function of LSTM, we can say that the LSTM updates its state by:

\begin{large}
\begin{equation} \label{eq2}
    h_t,c_t = f_{LSTM}(u_t,h_{t-1},c_{t-1})
\end{equation}
\end{large}
The internal computations of the LSTM, as described in Hochreiter et. al.\cite{lstm}, on its gates and memory cells are enumerated as follows:\\
\begin{large}
\begin{equation}\label{eq3}
i_t = \sigma (W_iu_t+ R_ih_{t-1}+ b_i)    
\end{equation}
\begin{equation}\label{eq4}
f_t = \sigma (W_fu_t+ R_fh_{t-1}+ b_f)    
\end{equation}
\begin{equation}\label{eq5}
o_t = \sigma (W_ou_t+ R_oh_{t-1}+ b_o)    
\end{equation}
\begin{equation}\label{eq6}
z_t = tanh (W_zu_t+ R_zh_{t-1}+ b_z)    
\end{equation}
\begin{equation}\label{eq7}
c_t = i_t\odot z_t + f_t \odot c_{t-1}     
\end{equation}
\begin{equation}\label{eq8}
h_t = o_t \odot tanh(c_t)   
\end{equation}
\end{large}

where \textit{W} denotes the input weight matrix learned during training, \textit{b} is the bias vector and \textit{R} is the recurrent weight matrix. Also, $\sigma$ denotes the sigmoid function which is expressed by $\sigma(x)=1/1+exp(x)$. It has a squashing effect and condenses the input into the range of (0, 1). $Tanh$ is hyperbolic tangent function and produces values in the range (-1,1) to avoid the explosive growth of values over time. Both the functions are computed in an element-wise manner.$i_t$, $o_t$ and $f_t$ denote the input, output and  forget gates respectively. To compute them we add the linear projections of $u_t$ and $h_{t-1}$ followed by the output of sigmoid function.The input transformation $z_t$, the cell value of the previous state $c_{t-1}$ and the output of the element-wise multiplication output which is denoted by $\odot$ are modulated by the input, forget and output gates, respectively. 

\subsection{Stacked LSTM}
\label{stacked}
We use a Three Layer stacked LSTM where the output of one layer is the input for the next and the output of the last layer is the output of the LSTM stack.
The initial hidden and cell states, $h^1_0$ and $c^1_0$ for the first LSTM layer, $h^2_0$ and $c^2_0$ for the second LSTM layer, $h^3_0$ and $c^3_0$ for the third LSTM layer are calculated by first calculating an average of the set of vectors $p$ defined in Equation \ref{eqp} and then feeding it through two separate Multi Layer Perceptrons represented as $f^i_{init\_c}$ and $f^i_{init\_h}$, $i= {1,2,3}$:
\begin{equation*}
    h^i_0 = f^i_{init\_h}(\dfrac{1}{L} \sum\limits_{i}^{L}a_i) \hspace{1cm}
    c^i_0 = f^i_{init\_c}(\dfrac{1}{L} \sum\limits_{i}^{L}a_i) 
\end{equation*}
where the annotation vectors $a_i, \ i = 1, . . . , L$  are the features that correspond to different image sub-regions.
Also, since we use a multi-layer LSTM stack, the first layer $LSTM^1$ takes word embeddings of the input sequence as the input as described in Section \ref{lstm_section}. For LSTM layers $LSTM^2$ and $LSTM^3$ the input is the output of the previous layer. 

\subsection{Encoder-Decoder Approach}
\label{inject}
In Encoder-Decoder Approach, the input is the sequence of word embeddings for the input word sequence generated at the last time step. At the first time step, the word embedding of the \begin{math} <start> \end{math} token is used as input to act as signal for the LSTM to start generating the output. Hence,
\begin{equation*}
    u_t = Ey_{t-1}
\end{equation*}
where $Ey_{t-1}$ is the sequence of word embeddings of the output at $t-1$ time-step.
\subsection{Soft Attention}
\label{softatten}
In the soft attention approach, in addition to the word embeddings of the input sequence, we also use the context vector $s_t$ which is a representation of image for that time-step which provides information about the relevant portion of the image for that time-step. Hence the LSTM outputs would be calculated according to the following equations:
\begin{large}
\begin{equation} \label{eq21}
    h_t,c_t = f_{LSTM}(u_t,z_t,h_{t-1},c_{t-1})
\end{equation}
\end{large}
where,
\begin{large}
\begin{equation}\label{eq31}
i_t = \sigma (W_iEy_{t-1}+ R_ih_{t-1}+ S_is_t+ b_i)    
\end{equation}
\begin{equation}\label{eq41}
f_t = \sigma (W_fEy_{t-1}+ R_fh_{t-1}+ S_fs_t+ b_f)    
\end{equation}
\begin{equation}\label{eq51}
o_t = \sigma (W_oEy_{t-1}+ R_oh_{t-1}+ S_cs_t+b_o)    
\end{equation}
\begin{equation}\label{eq61}
z_t = tanh (W_zEy_{t-1}+ R_zh_{t-1}+ S_cs_t +b_z)    
\end{equation}
\begin{equation}\label{eq71}
c_t = i_t\odot z_t + f_t \odot c_{t-1}     
\end{equation}
\begin{equation}\label{eq81}
h_t = o_t \odot tanh(c_t)   
\end{equation}
\end{large}

For the computation of the context vector $s_t$, we adopt the same mechanism as in Xu et. al.\cite{attendtell}, where for each annotation vector $a_i$ corresponding to a location in the image, a positive weight $\alpha_i$ is calculated by the attention mechanism which denotes the relative importance of the location for generating the word at the present time-step. The attention model $m_{att}$ is a function of $a_i$ and $h_{t-1}$, i.e, the annotation vector and the hidden state at time-step $t-1$.
\begin{equation*}
    x_{ti} = m_{att}(a_i, h_{t-1})
\end{equation*}
\begin{equation*}
    \alpha_{ti} = \dfrac{exp(e_{ti})}{\sum_{k=1}^{L}exp(e_{tk})}  
\end{equation*}
From the weights associated with each image region, the context vector $s_t$ can be calculated as 
\begin{equation}
    s_t = \Phi (\{a_i\},\{\alpha_i\})
\end{equation}
where $\Phi$ returns a single vector for the image.

\subsection{Data Augmentation}
\label{dataaug} We use random horizontal flipping and random perspective transformations for data augmentation. In random horizontal flipping, the image is flipped along the vertical axis. In other words, we take the mirror image of the image along the Y-axis on a plane. In random perspective transformation of the image, we change the orientation, and field of view of scene. Figure \ref{perspective_exp} shows some examples of perspective transformation applied to an example image (shown at the top left). During training, at each epoch we randomly apply both the transforms so that the model 'sees' different image at each epoch, thereby reducing the chances of overfitting. 

\begin{table}[h]
    \begin{tabular}{p{2.5cm}p{2.5cm}p{2.5cm}p{2.5cm}}
         \includegraphics[width=2.5cm,height=2.5cm]{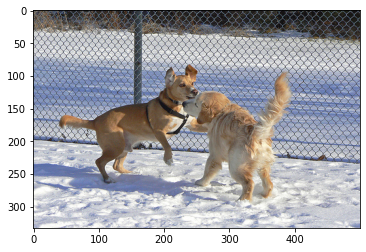}&
         \includegraphics[width=2.5cm,height=2.5cm]{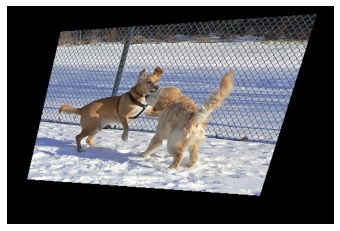}&
         \includegraphics[width=2.5cm,height=2.5cm]{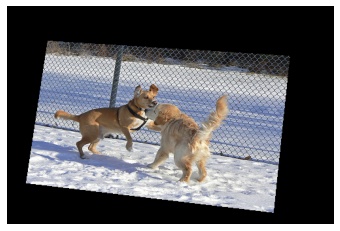}& 
         \includegraphics[width=2.5cm,height=2.5cm]{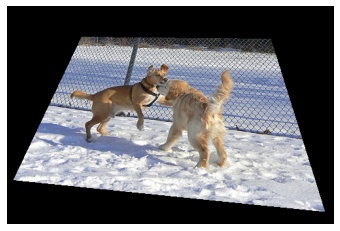}\\
         \includegraphics[width=2.5cm,height=2.5cm]{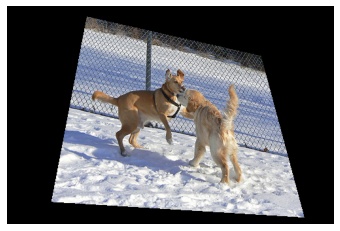}&
         \includegraphics[width=2.5cm,height=2.5cm]{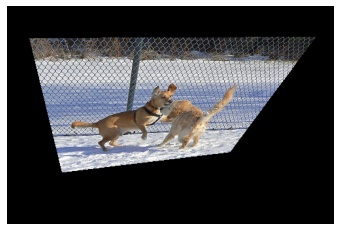}&
         \includegraphics[width=2.5cm,height=2.5cm]{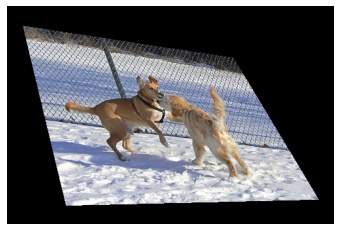}& 
         \includegraphics[width=2.5cm,height=2.5cm]{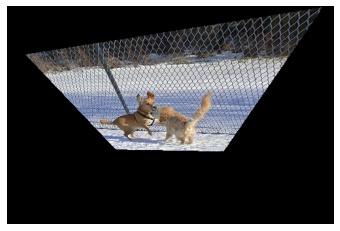}\\ 
    \end{tabular}
    \captionof{figure}{Some examples of perspective transformations of the input image at the top left.}
    \label{perspective_exp}
\end{table}

\section{Experiments and Results}
\label{experiments}
In this section we describe the Experimental setup, and the Results. We also discuss the Datasets used for training and evaluation, the metrics used for comparison with other methods proposed in the literature and the previous works in the literature that we have used for comparison with our methods.
\subsection{Datasets}
\label{datasets}
To measure the efficacy of our model, we have utilized the three widely datasets for image captioning: Flickr8k\cite{flickr8k}, Flickr30k\cite{flickr30k} and MSCOCO\cite{coco} datasets.\\
\textbf{Flickr8k} Dataset has around 8000 images with 6000 images reserved for training, 1000 images for validation and 1000 for testing. Each image has 5 captions for reference which have been annotated by humans.\\
\textbf{Flickr30k} has around 31000 images with each image having 5 human annotated captions for reference. Most images are about human activities that are observed in daily life. Following the work of Karpathy et. al. \cite{karpathy} and Xu et. al. \cite{attendtell}, we randomly select 1000 images for validation set, 1000 images for testing and the rest for training. 
The images for both Flickr8k and Flickr30k dataset have been collected from Flickr\footnote{www.flickr.com}.\\
\textbf{MSCOCO} dataset has around 164,000 images and unlike the Flickr8k and Flickr30k datasets, the official train, test and validation splits have been provided 82,783 images with their 413,915 human annotated captions for training, 40,504 images with their 202,520 human annotated captions for validation set and the rest for testing. It has been released by Chen et. al.\cite{chen}. However, for the test set the ground truth captions have not been provided. Hence, following the previous works (Mao et. al.\cite{mao}, Vinyals et. al.\cite{vinyals}, Xu et. al.\cite{attendtell}) we utilize the widely used train, test, validation splits provided in Karpathy et. al. \cite{karpathy}.  
\subsection{Evaluation Metrics}
\label{evaluation}
We evaluate our method on BLEU, METEOR and CIDEr metrics according to the methodology provided in MSCOCO evaluation\footnote{https://github.com/tylin/coco-caption} code provided with Chen et. al.\cite{chen} which also explains the metrics in detail. BLEU is a precision based metric and measures the fraction of words in the generated caption that also occur in the human-annotated references.

METEOR  computes alignment between the words in generated captions and the human annotated sentences to find a 1:1 correspondence. For calculation, in addition to matching the exact tokens, synonyms, stemmed tokens, paraphrases are also used. CIDEr metric measures a Term Frequency Inverse Document Frequency (TF-IDF) measure for each n-gram in the generated caption and reference sentence.

\subsection{Compared Methods}
\label{compared}
Here we describe, in brief, the methods that we have used for comparison with our model. We have selected the methods based on similarity of approach and also improvements the researchers have proposed over the existing mechanisms. Karpathy and Fei-Fei\cite{karpathy}  proposed a Multimodel Recurrent Neural Network Architecture where they used Region Convolutional Neural Network (RCNN) as encoder and a Bidirectional RNN, with a hidden layer size of 512 neurons, as decoder. Donahue et. al. \cite{donahue} propose a Long-term Recurrent Convolutional Networks (LRCNs) and use a multilayer decoder to process the information extracted from CNN in the first step. Vinyals et. al.\cite{vinyals}  propose an encoder decoder framework where the encoder is the Inception CNN and the decoder is an LSTM which has 512 neurons in the hidden layer. Mao et. al.\cite{mao} propose a model called Multimodal Recurrent Neural Network m-RNN where they use the AlexNet CNN to extract image features. They use two Word Embedding layers. The outputs of the second Word Embedding Layer, RNN, and the CNN are added together and used as input to the 512 dimensional Multimodal Layer to generate the caption. Xu et. al.\cite{attendtell} propose an attention based model where the decoder focuses on certain regions of the image at appropriate time-steps to generate captions that are more salient to the image. Another method is termed Part of Speech(POS) guidance in the work of He et. al.\cite{pos} where the POS tags of the current words are used as a cue to decide when to introduce image information to the LSTM decoder.
\subsection{Implementation Details}
\label{implementation}
We used workstations equipped with Quadro RTX 4000 GPUs and 32GBs of RAM. The implementation of the models was done using Python programming language utilizing the PyTorch Deep Learning Framework. The number of units in the LSTM layers was fixed at 512 and the batch size was 32. While using MSCOCO dataset, the training could be completed in less than 3 days whereas for Flickr30k dataset, it took about one and a half days.
\subsection{Quantitative Results}
\label{quantresults}
We compare our methods with the comparable methods proposed in the literature in Table \ref{results_all}.

\begin{table*}[h]
\scriptsize{
\begin{tabular}{p{3.0cm} p{1.1cm} p{1.1cm} p{1.1cm} p{1.1cm} p{1.3cm} p{1cm}}
\hline\noalign{\smallskip}
Method Name & BLEU-1 & BLEU-2  & BLEU-3 & BLEU-4 & METEOR & CIDEr \\
\noalign{\smallskip}\hline\noalign{\smallskip}
\noalign{\smallskip} 
\textbf{On Flickr8k} &&&&&& \\
Karpathy et al. \cite{karpathy} & 0.579 & 0.383 & 0.245 & 0.160 & \_\_ & \_\_  \\
Mao et al. \cite{mao} & 0.5778 & 0.2751  & 0.2307& \_\_  & \_\_  & \_\_  \\
Vinyals et al \cite{vinyals} & 0.63  & 0.41 &  0.27 & \_\_ & \_\_ & \_\_ \\
Soft-Attention \cite{attendtell} &0.67&0.448& 0.299 & 0.195 & 0.1893 & \_\_  \\
Hard-Attention  \cite{attendtell} &0.67&0.457&0.314&0.213 & 0.203 & \_\_ \\
\textit{Our Methods} &&&&&& \\
Encoder-Decoder &0.628&0.442&0.307&0.211&0.199&0.527 \\
Soft Attention &0.635&0.452&0.313&0.214&0.200&0.555 \\
\noalign{\smallskip}
\textbf{On Flickr30k} &&&&&& \\
Karpathy et al. \cite{karpathy} & 0.573& 0.369& 0.240& 0.157&\_\_&\_\_\\
Mao et al. \cite{mao} &0.5479& 0.2392& 0.1952& \_\_&\_\_&\_\_\\
LRCN \cite{donahue} & 0.5872& 0.3906& 0.2512& 0.1646& \_\_& \_\_\\
Vinyals et al \cite{vinyals} &0.663 &0.423&0.277 & 0.183 & \_\_& \_\_ \\
He et al \cite{pos} &0.638 &0.446 &0.307 &0.211& \_\_&\_\_ \\
Soft-Attention \cite{attendtell} &0.667& 0.434&0.288&0.191&0.1849&\_\_\\
Hard-Attention \cite{attendtell} &0.669 &0.439& 0.296& 0.199&0.1846&\_\_ \\
\textit{Our Methods} &&&&&& \\
Encoder-Decoder &0.655&0.468&0.329&0.229&0.188&0.482\\
Soft Attention &0.661&0.474&0.332&0.233&0.195&0.485\\
\noalign{\smallskip}
\textbf{On MSCOCO} &&&&&& \\
Karpathy et al. \cite{karpathy} & 0.625& 0.450& 0.321& 0.230& 0.195& 0.660\\
LRCN \cite{donahue} & 0.6686& 0.4892& 0.3489& 0.2492& \_\_& \_\_ \\
Vinyals et al \cite{vinyals} &0.666 &0.461 &0.329 &0.246&\_\_&\_\_\\
He et al \cite{pos} &0.711 &0.535 &0.388 &0.279 &0.239 &0.882\\
Soft-Attention \cite{attendtell} &0.707&0.492 &0.344&0.243 &0.239&\_\_\\
Hard-Attention \cite{attendtell}&0.718&0.504 &0.357 &0.250 &0.230&\_\_ \\
\textit{Our Methods} &&&&&& \\
Encoder-Decoder &0.720&0.551&0.409&0.302&0.254&0.998\\
Soft Attention &0.732&0.566&0.432&0.329&0.263&1.041 \\
\noalign{\smallskip}\hline
\end{tabular}
}
\caption{The performance comparison between our methods and the comparable models on Flickr8k, Flickr30k and MSCOCO datasets.The metrics for which the scores are unavailable in the cited works have been left blank.}
\label{results_all} 
\caption*{\footnotesize{}}
\end{table*}
\newpage

\begin{table}[h!]
    \centering
    \begin{tabular}{p{3cm}p{3cm}p{3cm}p{3cm}}
        (a)&(b)&(c)&(d)\\
         \includegraphics[height=3cm,width=3cm]
         {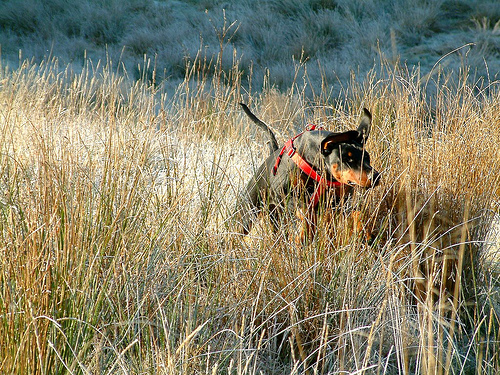}&
         \includegraphics[height=3cm,width=3cm]{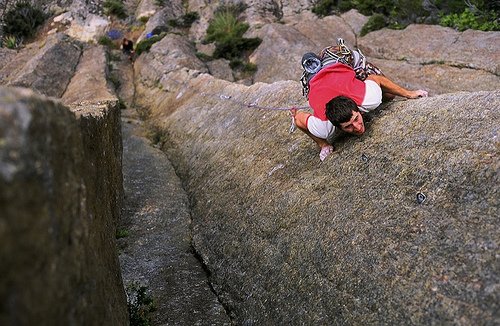}&
         \includegraphics[height=3cm,width=3cm]
         {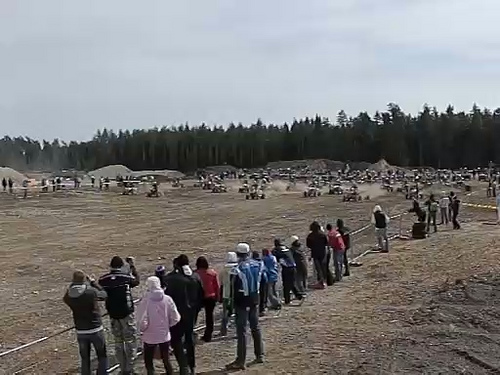}&
         \includegraphics[height=3cm,width=3cm]
         {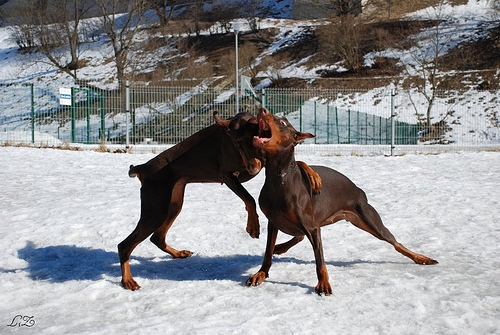}\\
         \small{ a black and white dog is running through the grass} &
         \small{a man in a red shirt is climbing climbing} &
         \small{a group of people in a line} &
         \small{two brown dogs are running through the snow} \\\hline
         \small{a dog with a red collar is running through tall grass} &
         \small{a person in a red shirt climbs a rock rock} &
         \small{a group of people are standing on a hill} &
         \small{two dogs are playing in the snow} \\
         (e)&(f)&(g)&(h)\\
         \includegraphics[height=3cm,width=3cm]{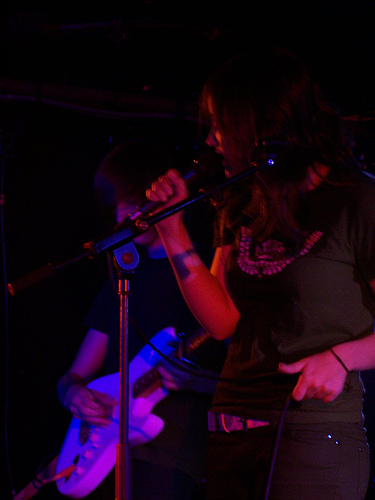}&
         \includegraphics[height=3cm,width=3cm]{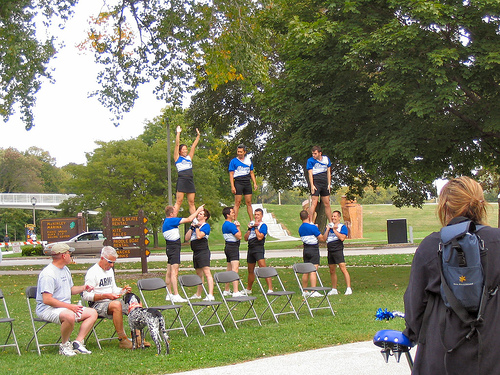}&
         \includegraphics[height=3cm,width=3cm]{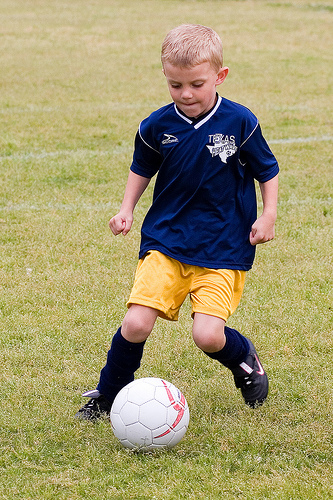}&
         \includegraphics[height=3cm,width=3cm]{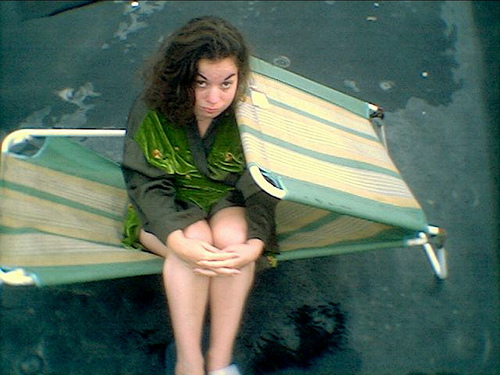}\\
         \small{a band is performing on stage}&
         \small{a group of people are gathered in the grass} &
         \small{a boy in a blue shirt and blue shorts is kicking a soccer ball} &
         \small{a woman in a black shirt is sitting on a bench} \\\hline
         \small{a man in a black shirt is holding a microphone} &
         \small{a group of people are sitting on a lawn} &
         \small{a little boy wearing a blue shirt kicking a soccer ball} & 
         \small{a woman in a green shirt is sitting on a boat} \\
         (i)&(j)&(k)&(l)\\
         \includegraphics[height=3cm,width=3cm]{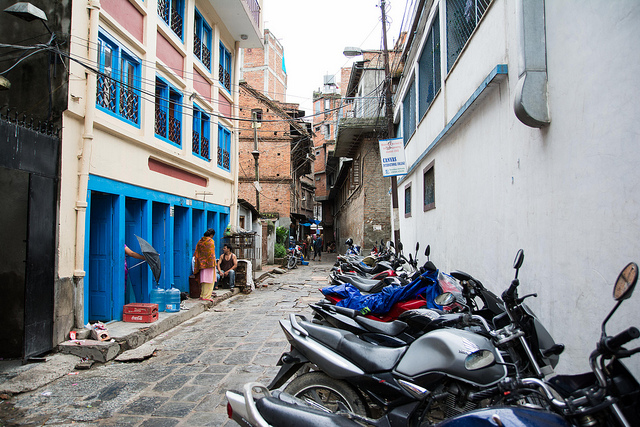}&
         \includegraphics[height=3cm,width=3cm]{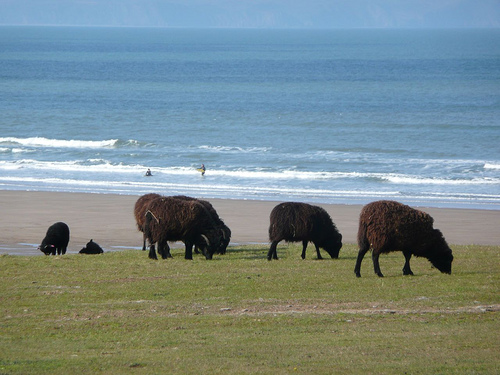}&
         \includegraphics[height=3cm,width=3cm]{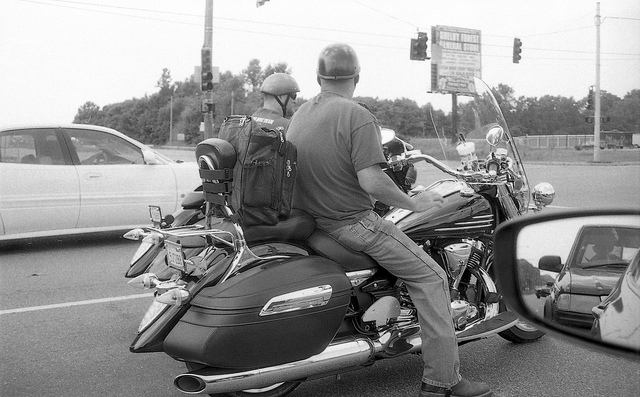}&
         \includegraphics[height=3cm,width=3cm]{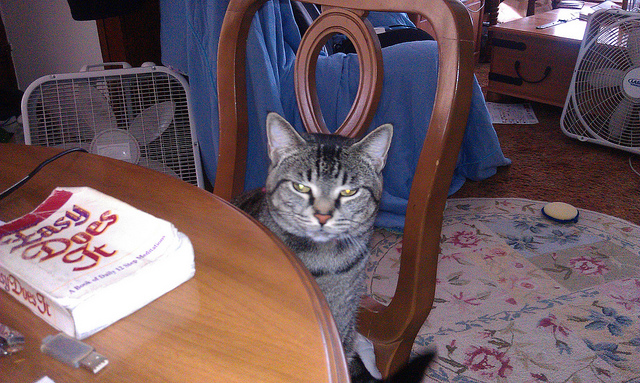}\\
         \small{a group of motorcycles parked in front of a building} &
         \small{a herd of sheep standing on top of a sandy beach} &
         \small{a man riding a motorcycle down a street} &
         \small{a cat sitting on top of a wooden table} \\\hline
         \small{a row of motorcycles parked next to a building} &
         \small{a herd of sheep grazing on a lush green field} &
         \small{a black and white photo of a man riding a motorcycle} &
         \small{a cat is sitting on a chair} \\
    \end{tabular}
    \captionof{figure}{Examples of generated captions. First,second and third rows have images from Flickr8k, Flickr30k and MSCOCO datasets respectively.}
    \label{gen_caps}
\end{table}

\begin{figure}
    \centering
    \begin{tabular}{c}
    \includegraphics[height=6.5cm,width=14cm]{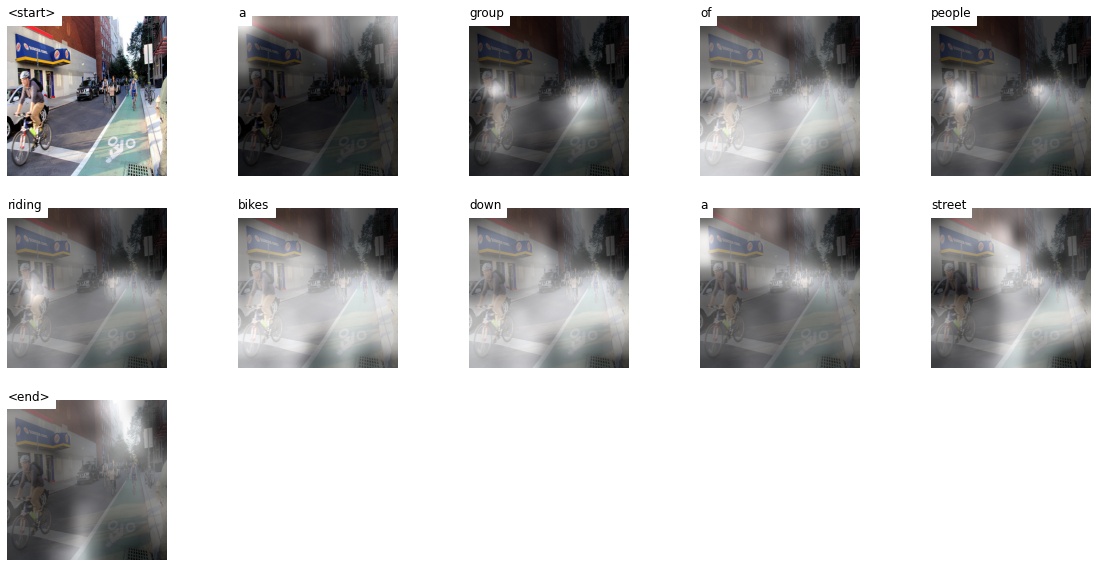}      \\
    (a)\\
    \includegraphics[height=6.5cm,width=14cm]{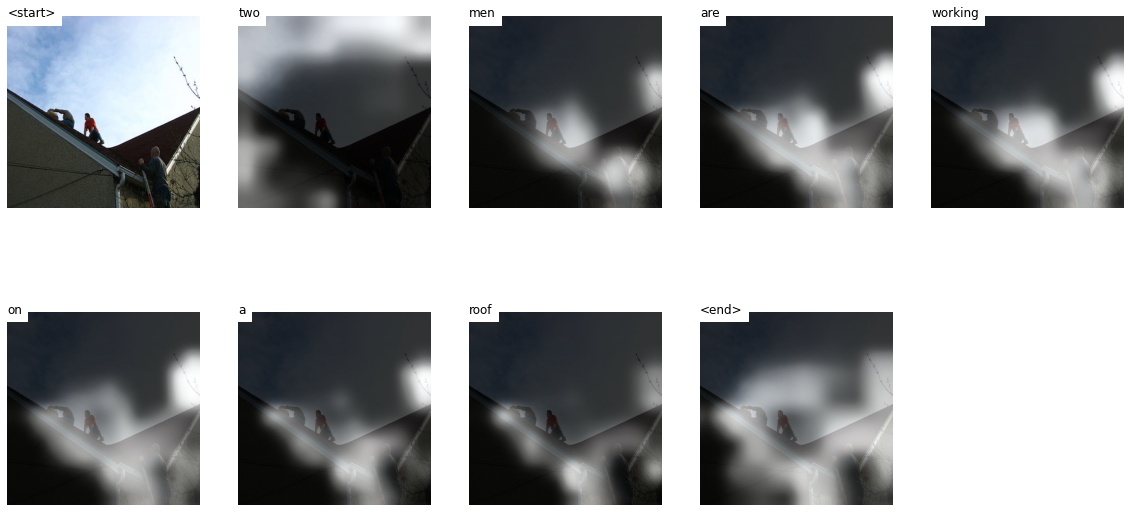}\\
    (b)
    \end{tabular}
    
    \caption{Visualization of soft attention at each time step. Figure (a) and (b) are for images from MSCOCO and Flickr30k datasets, respectively.}
    \label{att1}
\end{figure}

\subsection{Qualitative Results}
\label{qualitative}
We provide some examples of captions generated by our methods in Figure \ref{gen_caps}. For each image there are two generated captions: the upper row has captions generated by Encoder-Decoder Approach and the second row has captions generated by Soft Attention approach. As we can observe from the Figure \ref{gen_caps}, the Soft Attention approach is generally able to identify finer details such as color and smaller or obscure objects which the Encoder-Decoder Approach misses to identify. There are some outliers such as Figure \ref{gen_caps}(d) where the Soft Attention models doesn't mention the color although it performs activity recognition better than the Encoder-Decoder Approach Model. Figures \ref{gen_caps}(d), (h) and (l)  are examples where model didn't recognize the objects or the relationships among them correctly. In addition to correct interpretation of fine-grained details, the ability to focus on certain parts of the image does have other advantages as we can observe from Figure \ref{gen_caps}(k) where the Soft-Attention model is able to recognize that the picture is a grayscale image. However, we have observed the Soft-Attention model erroneously repeats the words (as in Figure \ref{gen_caps}(b)) more frequently than the
Encoder-Decoder Approach Model.

In Figure \ref{att1} we present two examples of attention visualization at each time step.

\section{Discussion and Conclusion}
If we compare the method proposed by Vinyals et al. \cite{vinyals} (named Vinyals et. al.\cite{vinyals} in Table \ref{results_all}) with the two Attention based Methods proposed by Xu et. al.\cite{attendtell} (named Soft Attention\cite{attendtell} and Hard Attention\cite{attendtell} in Table \ref{results_all}), we can see that attention based methods provide considerable improvements over the model without attention mechanism. However, our method which does not utilize Attention Mechanism (Encoder-Decoder Approach in Table \ref{results_all}) achieves comparable results with the method that uses Soft-Attention based approach. This is possibly because we use a better decoder with 3 LSTM layers stacked together and context based embeddings. Also, instead of extracting features from the last CNN layer we extract a set of feature vectors from intermediate Convolutional Layers.

We have presented a method to enhance the performance of Image Captioning using encoder-decoder approach. We have used three LSTM layers to allow the model to learn the complex relationships among words. And we have used contextual word embeddings to enable better utilization of relationships among words in the text. We have evaluated our approach with respect to two encoder-decoder frameworks: Encoder-Decoder Approach where the image features are provided at the first time-step and the Soft-Attention Approach where the model is provided a context vector at each time-step which allows the model to focus on certain parts of the image while generating a word. We have utilized InceptionResNet CNN to generate a set of feature vectors from the input image.

\section*{Acknowledgements}
No funding was received for this work from any source. However, we were generously allowed access to GPU-equipped workstations at the MultiMedia Processing Laboratory and the Language Processing Laboratory, both at the Department of Computer Science and Engineering, National Institute of Technology, Silchar for which we are extremely grateful.

\end{document}